\pdfoutput=1

\documentclass[11pt]{article}

\usepackage{acl}

\usepackage{times}
\usepackage{latexsym}

\usepackage[T1]{fontenc}

\usepackage[utf8]{inputenc}

\usepackage{microtype}

\usepackage{inconsolata}
\usepackage{amsfonts}
\usepackage{amsmath}

\usepackage[linesnumbered,ruled,vlined]{algorithm2e}
\SetAlgorithmName{Algo.}{Algolist}{List of Algorithms}
\SetKwComment{Comment}{/* }{ */}
\RestyleAlgo{ruled} 

\usepackage[most]{tcolorbox} 
\tcbuselibrary{skins}
\usepackage{xcolor} 
\usepackage{float}
\usepackage[framemethod=default]{mdframed}
\usepackage{multicol}
\mdfdefinestyle{exampledefault}{
    linecolor=blue,
    outerlinewidth=2pt,   
    roundcorner=50pt,      
    innertopmargin=5pt,
    innerbottommargin=5pt,
    innerrightmargin=5pt,
    innerleftmargin=0pt,  
    backgroundcolor=gray!10!white}
\definecolor{lightblue}{rgb}{0.8, 0.9, 1}

\usepackage{soul}
\newcommand{\hilight}[2][lightblue]{%
  \begingroup
  \sethlcolor{#1}
  \hl{\strut #2\strut}
  \endgroup
}

\usepackage{amsfonts}

\usepackage{textcomp}
\usepackage{booktabs}
\usepackage[inline]{enumitem} 
\usepackage{setspace} 
\usepackage{multirow} 
\usepackage{array}    
\newcolumntype{C}[1]{>{\centering\arraybackslash}p{#1}}
\usepackage{bm} 
\usepackage[T1]{fontenc} 
\newlist{inlineenum}{enumerate*}{1}
\setlist[inlineenum]{label=\textbf{\arabic*}.}
\title{Reinforcement Learning Problem Solving with Large Language Models}

\author{Sina Gholamian\\
  Thomson Reuters AI Labs\\
	Toronto, Canada\\
  \texttt{sina.gholamian@thomsonreuters.com} \\\And
  Domingo Huh\\
  Thomson Reuters AI Labs\\
  Toronto, Canada\\
  \texttt{domingo.huh@thomsonreuters.com} \\}

\begin{document}
\maketitle
\begin{abstract}
Large Language Models (LLMs) encapsulate an extensive amount of world knowledge, and this has enabled their application in various domains to improve the performance of a variety of Natural Language Processing (NLP) tasks. 
This has also facilitated a more accessible paradigm of conversation-based interactions between humans and AI systems to solve intended problems.
However, one interesting avenue that shows untapped potential is the use of LLMs as Reinforcement Learning (RL) agents to enable conversational RL problem solving.
Therefore, in this study, we explore the concept of formulating Markov Decision Process-based RL problems as LLM prompting tasks. 
We demonstrate how LLMs can be iteratively prompted to learn and optimize policies for specific RL tasks. 
In addition, we leverage the introduced prompting technique for episode simulation and Q-Learning, facilitated by LLMs. 
We then show the practicality of our approach through two detailed case studies for ``Research Scientist'' and ``Legal Matter Intake'' workflows.
\end{abstract}

\section{Introduction}
The recent inception of Generative Large Language Models into the limelight has significantly influenced and reshaped the applications of Natural Language Processing (NLP) and the ways that LLMs contribute to it~\cite{zhao2023survey}. 
Because LLMs are trained on a massive set of internet data, this comprehensive knowledge gives them the capabilities to extract semantics and respond coherently in a human-style conversation. 
Furthermore, fine-tuning through Reinforcement Learning with Human Feedback (RLHF)~\cite{ouyang2022training} has enabled LLMs to accurately follow input prompt instructions. 
These capabilities have allowed LLMs to be used for a variety of out-of-the-box applications, including chatbot agents~\cite{dan2023educhat,limna2023use}, coding~\cite{vaithilingam2022expectation,openai2023gpt4}, and general common-sense reasoning~\cite{wei2022chain}. 
In addition, some studies have shown that LLMs can be further leveraged for knowledge reasoning in professional domains such as in medical~\cite{lievin2022can} and legal~\cite{hakimi-parizi-etal-2023-comparative} fields.

In RL problems, one goal can be to optimize an objective function (e.g., the sum of the expected long-term rewards) as an agent interacts with the environment and receives feedback. 
Deep RL~\cite{mnih2015human}, a paradigm that implements the agent or policy with a deep learning model, has gained attention as a function approximator to implement RL agents and policies. 
Once the Deep RL model is trained, it can receive a set of observations as inputs, and output the corresponding action to navigate the agent to the next state. 
RL agents, and their policy implementations as RL models, do not necessarily have well-defined natural language interfaces, which limits their implementation accessibility to non-technical users. 
To this aim, prior work~\cite{woodward2020learning} has explored the advantages of expanding the RL agent interfaces to unstructured formats and has achieved improved task performance. 

In parallel, LLM-based prompting has introduced a new interaction paradigm between AI systems and end-users that is more intuitive and accessible to every-day and non-technical users. 
In addition, recent developments in prompting strategies, such as Chain-of-Thought (CoT) prompting~\cite{wei2022chain}, have improved LLMs' abilities to decompose complex tasks into manageable substeps and further enhance their reasoning and planning capabilities. 
This motivates the exploration of whether RL problems can be interactively solved through iterative prompting of the LLMs. 
Based on a set of input and output requirements for the RL task at hand, the LLM can arguably execute the task, evaluate whether the requirements are fully met, and determine if additional iterations are required to satisfy all the requirements. 
Subsequently, the interesting question of whether the embedded knowledge and reasoning capabilities of LLMs can be leveraged through iterative prompting to allow LLMs to function as RL agents remains unanswered. 

Consequently, in this research, we propose a new framework that leverages the reasoning and problem-solving capabilities of LLMs to align them for RL problem-solving. 
For this process, we leverage the latest paradigm of human and AI systems interactions, i.e., natural language prompting of LLMs. 
More specifically, we make the following contributions:
\begin{itemize}
\item We introduce an iterative prompting strategy for communicating with LLMs and formulate RL problems based on Markov Decision Process and Q-Learning into LLM prompting tasks. 
We then iteratively leverage the LLM's knowledge and self-reflection to optimize the RL problem and extract the desired output from the LLM. 
\item We propose the integration of episode generation and simulation into the prompting chain and thus enable LLM-based policy learning, and, subsequently, we solicit the optimal policy outcomes (i.e., episodes) from the LLM.
\item We provide two detailed case studies of how our approach can be effectively implemented to extract optimal policies for the Research Scientist and Legal Matter Intake workflows. 
\end{itemize}

\section{Prior Art}
While the primary applications of LLMs are in NLP tasks, and are receiving significant attention from the research community, the applications of LLMs in adjacent areas such as Reinforcement Learning (RL) have the potential to grow~\cite{wang2023survey,xi2023rise}. 
Previously, Janner et al.~\cite{janner2021offline} proposed an RL formulation via a sequence-to-sequence learning model based on the transformers architecture. 
However, this modeling lacks the more accessible natural language and prompt-based interactions with LLMs and human users. 
Consequently, recent studies have explored building RL agents with LLMs or enabling interactions between LLMs and RL agents to improve the accomplishment of specific tasks, to name a few, robot control~\cite{hu2023enabling} and game playing~\cite{xu2023language}. 

Prior work has shown prompt engineering scenarios can help in decomposing complex tasks and thus reach better outcomes with LLMs. 
Wei et al.~\cite{wei2022chain} introduced Chain-of-Thought (CoT) prompting, which prompts the LLM to solve a task in a step-by-step manner. 
Similarly, Tree-of-Thoughts (ToT)~\cite{yao2023tree} and Graph-of-Thoughts (GoT)~\cite{besta2023graph} expand on CoT and create tree and graph structures of prompts, such that different paths of the tree can break down the task into varied sets of smaller substeps. 
On an adjacent track, self-reflection prompting~\cite{shinn2023reflexion} aims to iteratively refine the output of the LLM and get closer to the desired output. 
Our approach is different from the aforementioned prompting methods as we introduce a framework for solving RL tasks with iterative prompting techniques that take into consideration the requirements of reaching an optimal RL policy.  

In essence, compared to prior approaches, we argue that our approach to successfully implementing natural language and prompt-based interactions with LLMs would be a more intuitive and potentially transformative method for interactions between human users and AI systems. 
Thus, it naturally represents the forthcoming paradigm for optimizing RL problems. 
\section{Preliminaries} 
An RL problem is often formally defined as a Markov Decision Process (MDP)~\cite{puterman2014markov}, where MDP provides a mathematical definition of the environment that the RL agent is interacting with. 
An MDP is defined as a tuple of $(S, A, P, R, \gamma)$, where $S = \{s_1, s_2, \ldots, s_n\}$ denotes a set of states that an agent can be within the environment; $A = \{a_1, a_2, \ldots, a_m\}$ represents a set of actions that agent can take at each state; $P$ is a state transition probability matrix that given the current state $s$ and action $a$ at time $t$, represents the probability of ending up in state $s'$ at the next time step, i.e., $P(s'|s, a) = Pr(S_{t+1} = s' \mid S_t = s, A_t = a)$; $R$ is a reward function that represents the expected reward for taking action $a$ in state $s$ and moving to state $s'$, $R(s, a, s') = \mathbb{E}[R_{t+1} \mid S_t = s, A_t = a, S_{t+1} = s']$;  $\gamma \in [0,1]$ is a discount factor that encourages more immediate rewards for $\gamma < 1$. 
Based on this framework we can define a policy ($\pi$) that is a strategy (i.e., function) that the agent can leverage to decide the next action to take based on the current state: $\pi(a|s) = Pr(A_t = a \mid S_t = s)$.

\subsection{Q-Learning}
One popular RL algorithm is Q-Learning which uses a Q-table to store the expected rewards for state-action pairs. 
The agent then uses this table to select the best action based on its current state. 
The Q-value update formula that chooses the action that maximizes the expected sum of rewards at time step $t$ is defined as:
\begin{align*}
Q(s_t, a_t) \leftarrow & Q(s_t, a_t) + \alpha \times \\& [r_{t+1} + \gamma \max_a Q(s_{t+1}, a) - Q(s_t, a_t)]
\end{align*} 
where $\alpha$ is the learning rate, $\gamma$ is the discount factor, $r_{t+1}$ is the reward received after taking action $a_t$ at state $s_t$, and $\max_a Q(s_{t+1}, a)$ represents the expected reward for the action that results in maximum reward at the next state. 
This algorithm can be augmented to balance both exploitation and exploration through $\epsilon$-greedy approach. 
With probability $p=\epsilon$, the agent takes a randomly chosen action from the set of all possible actions $A$ to encourage exploration, and otherwise selects the action that maximizes the reward, i.e., exploitation:\vspace{-2mm} 
\begin{equation*}
a_t = 
\begin{cases} 
a \sim \text{Uniform}(A) & : p = \epsilon \\
\underset{a}{\arg\max} \ Q(s_t, a) & :p = 1 - \epsilon 
\end{cases}
\end{equation*}\vspace{-3mm}
\section{Methodology}
We propose utilizing LLMs to formulate the RL problem as a prompting task and then leverage the internal knowledge of LLMs to train a Q-Learning RL agent, and thereby optimize the desired policy. 
Initially, it is necessary to clearly translate and define the communication of the requirements for an RL problem to an LLM in text-based prompts. 
Therefore, we define accurate steps to transform an RL problem into a set of prompts to elicit the desired outputs, in this case, the optimized policy, from the LLM.

\subsection{Prompt Framework} 
Based on the preliminaries discussed in the previous section for defining an RL problem within the MDP framework, we first outline a framework to provide the RL problem requirements through different sections of the prompt. 
Specifically, based on the tuple $(S, A, P, R, \gamma)$, we initially establish the context of the RL problem, followed by detailing states, actions, and rewards. 
We then compile these necessary steps in two elaborative case studies. 

\paragraph{Context Setup.} 
We define the problem context for the LLM, which sets the expected behavior of the LLM and outlines the inputs for the MDP-based RL problem. 
This context setup is crucial as it aligns the LLM to interpret and respond to the RL problem requirements.  

\begin{tcolorbox}[colback=blue!5!white,colframe=blue!50!white,bottom=0pt, top=1pt]
\textbf{Problem Context:} \textit{You are an RL agent tasked with maximizing cumulative reward for a given task. }\textit{You will be provided with the task, states, possible actions at each state, and rewards.}
\end{tcolorbox}

\paragraph{Task.}
We then specify the RL task that LLM is intended to optimize. 
The placeholder provided below indicated that it can be filled in with any arbitrary task.  
For example, the task targeted for optimization could be the \textit{``Workflow of a research scientist''}.
\begin{tcolorbox}[colback=blue!5!white,colframe=blue!50!white,bottom=0pt, top=1pt]
\textbf{Task:} \textit{\{place\_holder\}}
\end{tcolorbox}

\paragraph{Inputs.} 
Once the task is specified, we include the necessary inputs to formulate the task as a MDP-based RL problem. 
We specifically provide details on states, actions, and rewards. 
It is important to note that additional inputs such as $\gamma$ and $\epsilon$ can also be provided if required. 
If these parameters are not specified, the LLM will decide them. 
\begin{tcolorbox}[colback=blue!5!white,colframe=blue!50!white,bottom=0pt, top=1pt]
\textbf{States:} \textit{$\{s_1, s_2, \ldots, s_n\}$\\}
\textbf{Possible actions:} \textit{$\{a_1, a_2, \ldots, a_m\}$\\}
\textbf{Rewards:} \textit{$\{r_1, r_2, \ldots, r_k\}$}
\end{tcolorbox}

\paragraph{Requirements.}
This section can be added if specific task-related requirements are necessary. 
As a part of the requirements, we include episode simulation with the LLM.
RL agents require episodes of interactions with the environment to receive rewards and thereby evaluate different decisions. 
However, real-world interactions can often be time-consuming and costly. 
Hence, in line with our concept of utilizing LLMs as RL optimizers and agents, we also leverage LLMs as environment simulators. 
More specifically, in this setup, we request the LLM to generate episodes that begin from the start state and conclude at a terminal state. 
These episodes are then utilized for Q-Learning with the LLM. 
Below, we have listed some example requirements for Q-Learning and we provide  more detailed case studies in the later sections. 
\begin{tcolorbox}[colback=blue!5!white,colframe=blue!50!white,bottom=0pt]
\textbf{Requirements:} 
\begin{enumerate}[itemsep=-1ex]
\item Solve it with Q-Learning. 
\item Simulate the environment for $\{n\}$ episodes.
\item Episodes MUST begin at "Start" and finish at "End".
\item ...
\end{enumerate}
\end{tcolorbox}

\paragraph{Outputs.}
We then instruct the LLM regarding the expected output. 
We ask the LLM to return the Q table and values for the optimal episode. 
This way we can evaluate the effectiveness of our framework in achieving the optimal policy. 

\begin{tcolorbox}[colback=blue!5!white,colframe=blue!50!white,bottom=0pt, top=1pt]
\textbf{Output:} \textit{Print the Q-table, and list the state/action pairs and their "Q-values" for the optimal episode from "Start" to "End".}
\end{tcolorbox}

\paragraph{Iterative Check.} 
This step revisits the output of the LLM from the last iteration to confirm if it meets the requirements. 
Iterations can continue until the desired output or the maximum number of trials is reached. 
Similar approaches, like self-reflection~\cite{shinn2023reflexion}, have been shown to improve the alignment with the intended task and elicitation of desired outputs from LLMs.
\begin{tcolorbox}[colback=blue!5!white,colframe=blue!50!white,bottom=0pt, top=1pt]
\textit{Did your output satisfy all of the following requirements? If not, you MUST take a fresh approach and execute it if necessary.}\\\\
\{\textit{repeat} \textbf{``Task/States/Actions/Rewards''} and \textbf{``Requirements''} from \textit{above\}}
\end{tcolorbox}

\subsection{Algorithm} 
Algorithm~\ref{alg:steps} outlines the pseudocode for algorithmically implementing RL problem formulation to solve an arbitrary RL optimization problem. 
First, task-specific inputs are defined, followed by the crafting of a generic prompt incorporating these inputs,\textit{GeneratePrompt(.)} at Line 4. 
We then iterate in the while loop on Lines 6-11 to check whether the requirements are met or the maximum number of iterations is reached. 
If the LLM-generated output does not meet the requirements, the LLM is re-prompted with the Requirements U on Line 7. 
Finally, the output from the LLM, which should be the Q-table and the optimal episode for the learned policy according to the RL task, is returned.
This algorithm can be generally applied to any prompt-based LLM to iteratively find a solution to RL problems.
\vspace{-3mm}
\setlength{\textfloatsep}{2pt}
\begin{algorithm}
\caption{RL problem solving with LLMs}\label{alg:steps}
\KwIn{Problem Context $C$, Task $T$, States $S$, Actions $A$, Rewards $R$, Requirements $U$, and iterations $M$}
\KwOut{Q-Table for Policy $\pi$}
$Satisfied \gets False$\;
$Max\_Iter \gets M$\;
$iter \gets 0$\;
$Prompt \gets GeneratePrompt(C,T,S,A,R,U)$\Comment*[r]{e.g., Figure~\ref{prompt_all}}
$Output \gets LLM(Prompt)$\;
$Prompt \gets GeneratePrompt(T,S,A,R,U)$\Comment*[r]{e.g., Figure~\ref{prompt_iterative}}
\While{$iter < Max\_Iter$}{
	$Satisfied \gets LLM(Prompt, Output)$\;
  \eIf{$Satisfied$}{
  \KwRet $Output$\; 
  }{
   $Output \gets LLM(Prompt)$\;
   $iter \gets iter + 1$ 
   }
}
\KwRet $Output$\;
\end{algorithm}
\vspace{-6mm}
\section{Case Studies}
In this section, we provide two illustrative case studies on how our approach can be applied to optimize different workflows.
\subsection{Research Scientist Workflow}  
In this case study, we consider our task to be optimizing the workflow of a Research Scientist.
The professional is involved in daily research activities like literature reviews and research publications. 
The goal is to learn and optimize the sequence of steps that a researcher should take to optimize their workflow. 

To address this problem with Q-Learning through LLM agents, we first create a simulation of the environment based on the provided input states, actions, and rewards. 
Then, the Q-Learning algorithm is implemented and iterated for a predefined number of episodes (e.g., 1000) to optimize the policy. 
Each episode begins from the initial state, i.e., `Start', and ends at the terminal state, i.e., 'End'. 
To ensure convergence, other parameters, such as the maximum number of steps per episode can also be defined.  
After training, the leaned policy is stored in the Q-table and we can identify and extract episodes with the highest cumulative rewards.
As an example, we provide the following states, actions, and rewards to be included in the LLM's prompt. 
For simplicity, we assume equal probabilities for existing transitions from one state to another. 
Later, we will discuss how this condition can be relaxed by deriving the probabilities from actual workflows. 
\begin{tcolorbox}[colback=gray!10!white, colframe=gray!75!black,arc=2mm,boxsep=1pt]
\begin{flushleft}
\textbf{States} = \{`Start (ST)', `Initiate Research (IR)', `Literature Review (LR)’, `Experiment Plan (EP)', `Experiment Execution (EE)', `Data Analysis (DA)', `Manuscript Drafting (MD)', `Submission to Venue (SV)', `Revision of Manuscript (RM)', `Peer Review (PR)', `Result Publication (RP)', `End (ED)'\}

\end{flushleft}
\end{tcolorbox}

\begin{figure}[htbp]
\vspace{-5mm}
  \centering
  \includegraphics[width=1\linewidth]{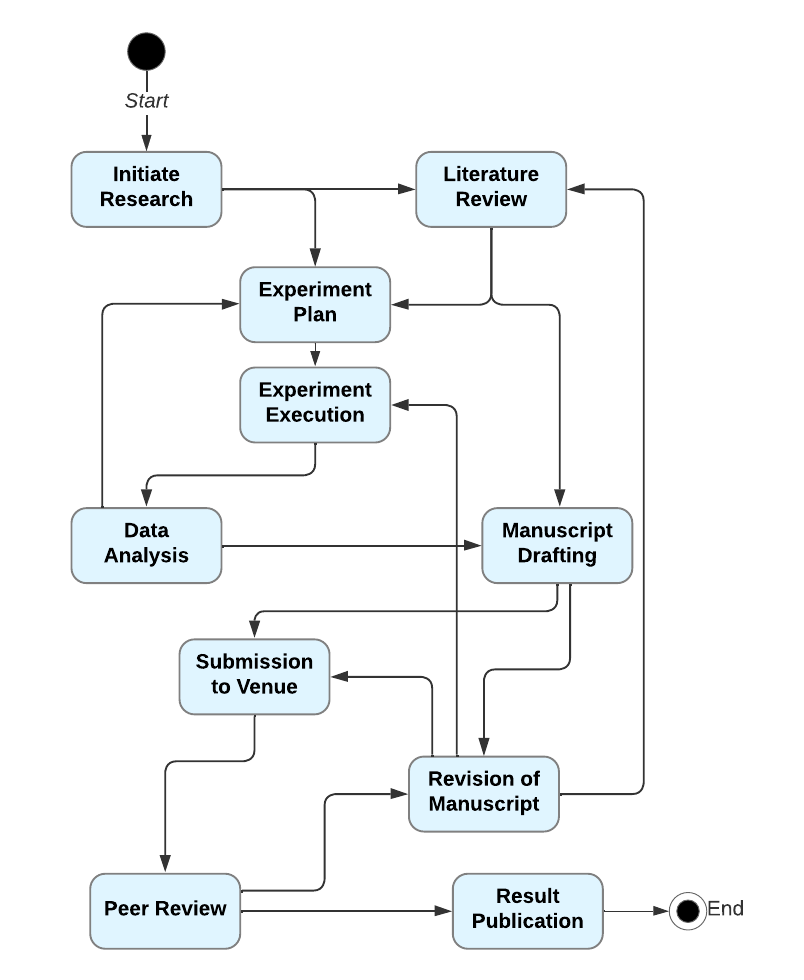}
  \captionsetup{skip=-3pt}
  \caption{\small Research Scientist workflow.}
  \label{rs_flow}
  \vspace{-2mm}
\end{figure}

Similarly, we define the possible actions for each state as follows. 
For every state $s$ we list all possible actions $a$ that can be taken from that state, $\forall s \in S| \text{ $s$: }[a_{p1},\ldots, a_{pn}]$, and for simplicity, we assume that actions from each state initially have the same probability. 
States and action combined define the state machine for the MDP. 
Figure~\ref{rs_flow} depicts the state machine for the research scientist workflow. 
While we have aimed to represent a meaning flow in this case study, it is for demonstration purposes and may not necessarily fully generalize to real-life scenarios. 

\begin{tcolorbox}[colback=gray!10!white, colframe=gray!75!black,arc=2mm,boxsep=-2pt]
\begin{flushleft}
\textbf{Actions} = \{`ST': [`IR'], `IR': [`LR', `EP'], `LR': [`EP', `MD'], `EP': [`EE'], `EE': [`DA'], `DA': [`MD', `EP'], `MD': [`SV', `RM'], `SV': [`RM', `PR'], `RM': [`EE', `SV'], `PR': [`RM', `RP'], `RP': [`ED'], `ED': [`ED'], `ELSE': -inf\}

\end{flushleft}
\end{tcolorbox}

Additionally, we assign rewards accordingly to optimize the workflow. 
In this scenario, we aim to reach the `End', i.e., the terminal state, as soon as possible. 
Hence, for simplicity, we assign a reward of $-1$ to every state, except the terminal state, which receives a reward of $0$. 
\begin{equation*}
r(s) = 
\begin{cases} 
    -1 & \text{if } s \neq \text{`End (ED)'} \\
    0 & \text{if } s = \text{`End (ED)'} 
\end{cases}
\end{equation*}

\begin{tcolorbox}[colback=gray!10!white, colframe=gray!75!black,arc=2mm,boxsep=-2pt]
\begin{flushleft}
\textbf{Rewards} = \{`ED': 0, else: -1\}
\end{flushleft}
\end{tcolorbox}
Lastly, we define the requirements for optimizing the task of ``Research Scientist Workflow'' as follows. 

\begin{tcolorbox}[colback=gray!10!white, colframe=gray!75!black,arc=2mm,boxsep=1pt,left=2pt,right=2pt,top=2pt,bottom=0pt]
\begin{flushleft}
\begin{spacing}{1}
\textbf{Requirements:}
\begin{enumerate}[itemsep=-1ex]
\item Solve it with Q-Learning. 
\item First, simulate the environment for 1000 episodes and fill in the Q-table.
\item Episodes MUST begin at "Start" and finish at "End".
\item ...\vspace{-5mm}
\end{enumerate}
\end{spacing}
\end{flushleft}
\end{tcolorbox}

The requirements ensure that the LLM initially simulates the MDP for a given number of episodes and uses Q-Learning. 
This method could easily generalize to a new set of requirements and algorithms, e.g., SARSA instead of Q-Learning. 
Finally, we prompt the LLM to provide the output and iteratively check whether it satisfies the requirements.

\subsection{Legal Matter Intake Workflow}\vspace{-1mm}  
As the second case study, we illustrate the workflow for tracking legal matters. 
This flow outlines the steps that legal associates within law firms typically follow when receiving a new legal request from a client and managing it through to the completion of the matter. 
The flow generally consists of the following steps: 

\begin{inlineenum}
\item \textbf{Matter Intake (MI):} The first step involves taking in the client's request and assessing whether the firm has the required expertise and resources to undertake this matter. 

\item \textbf{Conflict Assessment (CA):} Upon intake, the firm or attorney involved is required to assess conflicts of interest to ensure they are not precluded from representing the potential client. 
In some cases, CA can be bypassed, for example, for existing clients.

\item \textbf{Initial Assessment (IA):} An associate within the legal firm then conducts an initial assessment to identify client's case strengths and weaknesses. 
The associate will also come up with a plan on how to proceed with the case. 
This step may circle back to CA if further information are required after the initial assessment.  

\item \textbf{Client Communication (CC):} In this step, the legal firm communicates with the potential client, briefing them on the outcomes of the initial assessment and providing an overview of the next legal steps. 
This also give an opportunity to the client to discuss any questions they may have. 

\item \textbf{Fee and Payment (FP):} This step involves communicating the legal fees or estimations associated with the case to the client, such as hourly rates or any contingency fees. 
This step can be potentially bypassed assuming the client is already aware of the law firm's fees. 

\item \textbf{Proposal Preparation (PP):} Based on the initial assessment and the client's needs, the legal professional prepares a proposal that outlines the scope of work, the estimated cost, and the expected outcome.

\item \textbf{Proposal review (PR):} Once the proposal is prepared, the legal professional reviews it with the client and address any questions or concerns. 
After the client agrees with the plan, the matter progresses to the Case Management step. 

\item \textbf{Case Management (CM):} This step contains the bulk of the legal research work. 
CM entails that the legal professional will conduct legal research, draft legal documents, and potentially represent their client in court or other legal proceedings.

\item \textbf{Billing (BI):} At last, the law firm manages the billing and invoicing process, which includes tracking the time spent on the case, i.e., billable hours, preparing invoices, and sending them to the client for payment.
\end{inlineenum}

\begin{figure}[htbp]
\vspace{-4mm}
  \centering
  \includegraphics[width=1\linewidth]{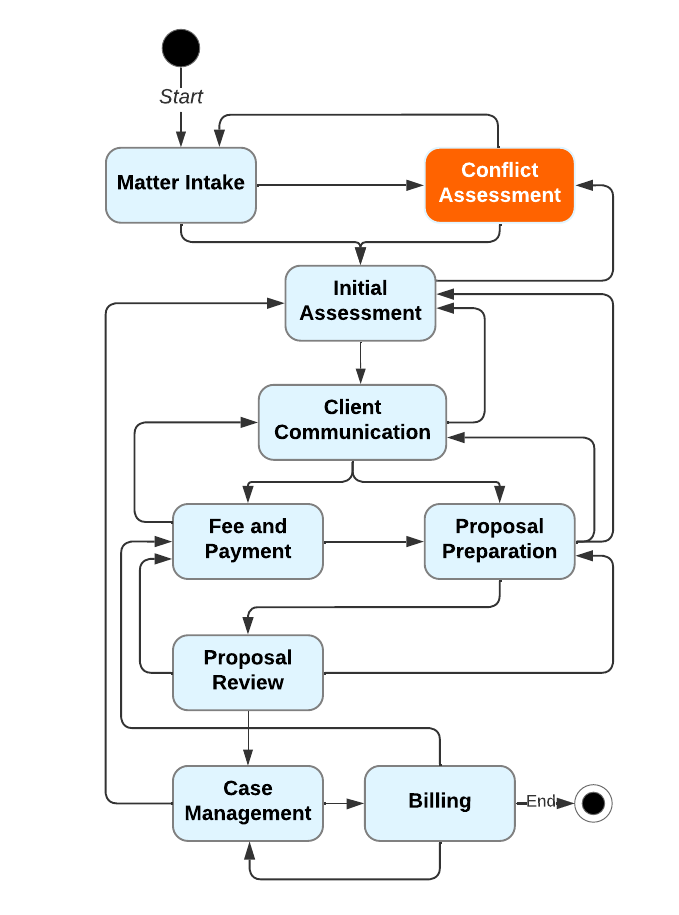}
  \captionsetup{skip=-2pt}
  \caption{\small Legal Matter Intake workflow.}
  \label{lm_flow}
  \vspace{2mm}
\end{figure}

Based on the above-mentioned possible stages of the legal matter intake process, we produce a feasible arrangement of states and potential transitions among them in Figure~\ref{lm_flow}. 
It is important to reiterate that this workflow is intended for demonstration purposes only and might not accurately capture the potential variability of this process. 
Similar to the research scientist workflow, and based on Figure~\ref{lm_flow}, \textbf{States} and \textbf{Actions} for legal matter intake workflow are outlined below.  
\begin{tcolorbox}[colback=gray!10!white, colframe=gray!75!black,arc=2mm,boxsep=1pt]
\begin{flushleft}
\textbf{States} = \{`Start (ST)', `Matter Intake (MI)', `Conflict Assessment (CA)', `Initial Assessment (IA)', `Client Communication (CC)', `Fee and Payment (FP)', `Proposal Preparation (PP)', `Proposal Review (PR)', `Case Management (CM)', `Billing (BI)', `End (ED)'\}
\end{flushleft}
\end{tcolorbox}
\begin{tcolorbox}[colback=gray!10!white, colframe=gray!75!black,arc=2mm,boxsep=-2pt]
\begin{flushleft}
\textbf{Actions} = \{`ST': [`MI'], `MI': [`CA', `IA'], `CA': [`IA', `MI'], `IA': [`CC', `CA'], `CC': [`FP', `IA', `PP'], `FP': [`PP', `CC'], `PP': [`PR', `CC', `IA'], `PR': [`CM', `PP', `FP'], `CM': [`BI', `IA'], `BI': [`CM', `ED', `FP'], `ED': [`ED'], `ELSE': -inf\}
\end{flushleft}
\end{tcolorbox}
For each state, possible transitions to other states are listed as valid actions. 
At last, we include a provision in our framework to inform the LLM that the remaining transitions are invalid, using a special notation at the end of the list of actions: \textbf{`ELSE': -inf}. 
This approach allows the Q-Learning algorithm to model invalid transitions with a infinitely large negative reward. 
As a result, in order to maximize cumulative reward, our framework, in principle, avoids these actions. 
For legal matter intake workflow, \textbf{Rewards}, \textbf{Requirements}, and the details of episode simulation closely follow the discussion mentioned for research scientist workflow, as such for brevity, we avoid repeating them here.

\section{Evaluation}\vspace{-1mm}
We experimented with our prompting framework with the GPT-4 model (knowledge cutoff: April 2023) for both case studies which includes four configurations,  and for 10 iterations per configuration. 
In cases that the GPT model initially fails to deliver the correct answer, we re-prompt it with{\setstretch{1.5} \textit{\hilight{``Did your output satisfy all of the following requirements? If not, you MUST take a fresh approach and execute it if necessary.''}}}, and we repeat the states, actions, rewards, and requirements (refer to Figure~\ref{prompt_iterative} in the appendices), with a maximum of 5 iterations, $Max\_Iter=5$. 
Table~\ref{workflow_table} summarizes the distribution for the number of iterations and the expected reward for the optimal workflow for both research scientist and legal matter intake. We also test two $\gamma$ (discount factor) configurations. 
We first do not specify (UNS) the $\gamma$ value and allow the LLM to decide its own value which can be variable on each iteration. 
In the second configuration, we provide a fix value for $\gamma$ as a part of the task requirements and then measure the variability of the calculated optimal reward for both cases. 
Overall, we run the experiments for 40 iterations, 20 iterations for each use case.    

\subsection{Research Scientist}
In 19 out of 20 iterations for this use case, our approach resulted in the following optimal workflow for the research scientist: \textit{\hilight{Start (ST) \textrightarrow Initiate Research (IR) \textrightarrow Literature Review (LR) \textrightarrow Manuscript Drafting (MD) \textrightarrow Submission to Venue (SV) \textrightarrow Peer Review (PV) \textrightarrow Result Publication (RP) \textrightarrow End (ED)}}. 
This sequence is optimal for this use case and bypasses common workflow steps such as experiment planning and execution. 
This outcome confirms that the LLM is not outputting a commonly followed flow based on its embedded knowledge if not optimal. 
The optimal reward considering  $\gamma=0.9$ is $-4.7$. 
Variations in the LLM's calculation of the expected reward for UNS configuration are generally attributed to the use of different $\gamma$ values, and we observe that setting $\gamma$ value as part of the input requirements results in zero reward variations.    

\subsection{Legal Matter Intake}
Similarly, for legal matter intake workflow, in all 20 iterations, the LLM was able to reach the optimal flow: \textit{\hilight{Start (ST) \textrightarrow Matter Intake (MI) \textrightarrow Initial Assessment (IA) \textrightarrow Client Communication (CC) \textrightarrow Proposal Preparation (PP) \textrightarrow Proposal Review (PR) \textrightarrow Case Management (CM) \textrightarrow Billing (BI) \textrightarrow End (ED)}}. 
The optimal reward considering  $\gamma=0.9$ is $-5.2$, and similar to the first use case, setting the $\gamma$ value as part of workflow requirements results in zero variance in optimal values.   

\begin{table}[h]
\centering
\small
\begin{tabular}{C{2.35cm}|C{.5cm}|C{1.2cm}|C{2cm}}
\toprule
\textbf{Task} & \textbf{($\bm{\gamma}$)} & \parbox{11mm}{\centering \textbf{Iterations \\ $\mathcal{N}(\mu, \sigma)$}}  & \parbox{23mm}{\centering \textbf{Optimal Reward}  \\ $\mathcal{N}(\mu, \sigma)$} \\
\midrule
\multirow{2}{*}{Research Scientist}   &UNS& $(1.6, 0.5)$   & $(-4.3, 1.8)$  \\
 & $0.9$ & $(2.0, 0.5)$  & $(-4.7, 0.0)$  \\ 

\midrule
\multirow{2}{*}{Legal Matter Intake}   & UNS & $(1.8, 0.4)$   & $(-4.7, 2.0)$  \\
 &$0.9$ & $(1.5, 0.5)$ & $(-5.2,0.0)$ \\

\bottomrule
\end{tabular}
\caption{Results for Research Scientist and Legal Matter Intake workflows. UNS: Unspecified $\gamma$ value.}
\label{workflow_table}
\vspace*{-1mm}
\end{table}

For both use cases, on average, our framework was able to reach the optimal workflows with $\le 2$ iterations, as illustrated on \textbf{"Iterations"} distribution in Table~\ref{workflow_table}. 
Our observation was that in most cases, LLM was able to satisfy most of the requirements in solving the task optimization in the first prompt, i.e., Figure~\ref{prompt_all}. 
The second prompt, i.e., Figure ~\ref{prompt_iterative}, then allowed the LLM to revisit the requirements and fix if some of the conditions are missed in the first trial. 
Overall, our approach demonstrated the effectiveness of iterative prompting in order to ensure that LLM is navigated towards satisfying all task optimization requirements.  

\section{Discussion}
\subsection{RL Task Complexity}
Although our provided use cases might seem limited, we believe as the available prompt context of LLMs is increasing over 100K tokens ~\cite{OpenAI2023DevDayAnnouncements,Anthropic2023Claude2.1}, the potential to handle more complex RL problems using our proposed framework is growing. 
Additionally, with the recent announcement of Custom GPTs~\cite{OpenAI2023IntroducingGPTs}, we anticipate that LLMs that are tailored to specific user preferences and requirements will further allow us to optimize according to our RL framework. 
Therefore, our approach can be integrated as a custom NLP module that customizes an LLM for optimizing RL tasks, while allowing the end user to interact with LLM-based agent through natural language.\vspace{-1mm} 
\subsection{User Logs}\vspace*{-1mm}
In our research scientist and leggal matter intake examples, we demonstrated how a workflow can be optimized as an RL problem. 
This approach can be applied to workflow logs from various personnel in an enterprise setting instead of the simulated workflows that we generated with the LLM. 
By analyzing these workflow logs and targeting metrics to optimize, such as task efficiency, we can optimize the workflows of employees, and thus improve overall enterprise efficiency. 
Additionally, we initially relied on the LLM to generate simulations and episodes of interactions with provided states, actions, and rewards. 
Compared to the LLM simulations, the available user log data can better represent the actions and rewards for different states and result in a more accurate implementation of our MDP framework. This aspect serves as an interesting avenue for our future research.

\subsection{LLM-based Planning}
In our work, we primarily aimed to describe the plan of solving the use case for LLM to follow. 
An extension of our work to consider is to perform this step with the LLM as well~\cite{valmeekam2023on}. 
Such that we first ask the LLM to generate several plans to solve the use cases and list down the steps involved in each. 
We then navigate the LLM through the steps for each plan via our prompting platform and collect the outcomes of different plans. 
Lastly, the LLM can be also serve as a judge to select the best plan based on the outcomes of different plans. 

\begin{figure}[t]
\vspace{-6mm}
  \centering
  \includegraphics[width=1\linewidth]{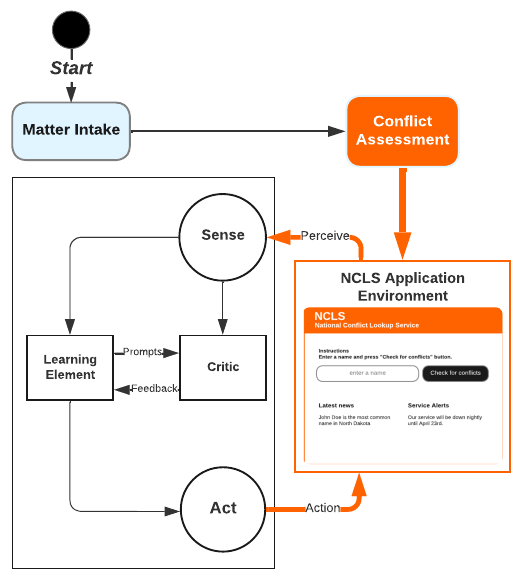}
  \caption{\small MLLM agent for Conflict Assessment state.}
  \label{lm_agent}
  \vspace{2mm}
\end{figure}

Furthermore, every step of the planning can also be delegated to specialized multimodal LLM agents (MLLMs)~\cite{yin2023survey}. 
To provide a concrete example, we dive into the "Conflict Assessment" in the legal matter intake flow as illustrate in Figure~\ref{lm_agent}. 
Assume the agent needs to check for conflicts through an interactive application called National Conflict Lookup Service (NCLS). 
The information captured during the "Matter Intake" state includes the client's full name along with case-related details and is available for use by the "Conflict Assessment" MLLM. 
The specialized MLLM has learned how to successfully perceive the NCLS application and act on it by entering the client's full name into the appropriate text field, pressing the "Check for conflicts" button, and resulting in either a "No conflict found" or "Conflict found" status update, so it can transition to the next ideal workflow state. 

The left box on Figure~\ref{lm_agent} provides a sketch on how the MLLM agent can learn to complete the task through interactions with the NCLS environment, similar to ScreenAgent~\cite{niu2024screenagent}. 
Upon entering the "Conflict Assessment" state, and with an interactive RL-based problem-solving approach in mind, the MLLM agent observes the state's default prompt for the task, e.g., "Check client's name for conflict". 
Assuming a first pass at performing the task, the \textit{Learning Element} \textit{Acts} on the NCLS and \textit{Senses} the observations. 
The \textit{Critic} provides feedback on the agent's actions and allow it to refine its approach to performing the task. 
In essence, the MLLM agent will learn and iteratively rephrase the current state's prompt in "Check for conflicts". 
We pursue MLLM-based planning and task optimization as directions for our future work. 

\section{Conclusion}\vspace*{-1mm}
With the advent of Large Language Models (LLMs), recent research has aimed to build on their reasoning and problem-solving capabilities and adapt them to different complex tasks. 
In this study, we introduced a novel approach to frame reinforcement learning problems as LLM prompting tasks as a novel way of solving RL problems through the user and LLM interactions. 
We developed an LLM prompting methodology that represents different elements of an RL problem based on the Markov Decision Process. 
Additionally, we leveraged LLMs for episode simulation and Q-Learning optimization and derived the optimal policy and reward based on our iterative prompting framework. 
In future work, we plan to expand our work to additional tasks and optimize workflow processes in the enterprise environment with LLM-based planning.  

\section{Limitations}
\paragraph{Scope.}
Although we aimed to introduce a generic approach for RL problem formulation, a few limited cases are experimented with. 
Thus, while we think our prompting technique should be generalizable to other RL tasks, further experimentation to verify broader evidence is required. 
\paragraph{LLMs Variability.}
Different LLMs may have noticeable differences in their performance on our illustrated use cases, and thus, can impact our findings. 
In addition, in our use cases, we also observe that LLM outputs for different iterations can be variable and it can often be difficult to reproduce the same exact results through prompting. 
For this reason, we conducted our study on a state-of-the-art (SOTA) LLM to establish the initial viability of our framework. 
We believe that as the LLMs continue to improve and as SOTA advances, this only further validates the feasibility of expanding our RL framework to tackle more complex RL or planning problems. 
Furthermore, to achieve more consistent LLM outputs and structured data formats, approaches such as LLM Functions~\cite{OpenAI2023FunctionCalling} can be leveraged.
\paragraph{Ethical Considerations.} 
As we use LLMs as the basis of our work, ethical considerations evolve on the appropriateness of the generated outputs~\cite{head2023large}. 
We reason that the chance of harmful outputs for our application should be minimal as our framework aligns the prompts to specific RL tasks. 
Furthurmore, when analyzing real user workflow logs, it is imperative to implement steps to ensure privacy.
\bibliography{agents_refs}

\appendix
\begin{figure*}[!ht]
\begin{tcolorbox}[left=1pt,right=1pt,colback=blue!5!white,colframe=blue!50!white]
\textbf{Problem Context:} You are an RL agent tasked with maximizing cumulative reward for a given task.\\

You will be provided with the task, states, possible actions at each state, and rewards.\\

\textbf{Task:} Workflow of a research scientist.\\

\textbf{States} = \{`Start (ST)', `Initiate Research (IR)', `Literature Review (LR)’, `Experiment Plan (EP)', `Experiment Execution (EE)', `Data Analysis (DA)', `Manuscript Drafting (MD)', `Submission to Venue (SV)', `Revision of Manuscript (RM)', `Peer Review (PR)', `Result Publication (RP)', `End (ED)'\}\\

\textbf{Actions} = \{`ST': [`IR'], `IR': [`LR', `EP'], `LR': [`EP', `MD'], `EP': [`EE'], `EE': [`DA'], `DA': [`MD', `EP'], `MD': [`SV', `RM'], `SV': [`RM', `PR'], `RM': [`EE', `SV'], `PR': [`RM', `RP'], `RP': [`ED'], `ED': [`ED'], `ELSE': -inf\}\\

\textbf{Rewards} = \{`ED': 0, else: -1\}\\

\textbf{Requirements:}\vspace*{-2mm}
\begin{enumerate}[itemsep=-1pt]
\item Solve it with Q-learning. gamma=0.9
\item Transitions from one state to another are only allowed based on provided 'Actions'. Other actions are not possible.
\item First, simulate the environment for 1000 episodes and fill in the Q-table for states. If an episode goes more than 100 steps terminate that episode.
\item Episodes MUST begin at "Start" and finish at "End".
\end{enumerate} \vspace*{1mm}

\textbf{Output:}\\
Print the Q-table, and list the state/action pairs and their "Q-values" for the optimal episode from "Start" to "End".
\end{tcolorbox}\vspace{-3mm}
\caption{This prompt is used to model the research scientist workflow based on the Markov Decision Process.}
\label{prompt_all}
\end{figure*}

\begin{figure*}[!ht]
\begin{tcolorbox}[left=1pt,right=1pt,colback=blue!5!white,colframe=blue!50!white]
\textbf{Did your output satisfy all of the following requirements? If not, you MUST take a fresh approach and execute it if necessary.\\}

\textbf{{Task:}} Workflow of a research scientist.\\
\textbf{States:} \textit{$\{s_1, s_2, \ldots, s_n\}$\\}
\textbf{Actions:} \textit{$\{a_1, a_2, \ldots, a_m\}$\\}
\textbf{Rewards:} \textit{$\{r_1, r_2, \ldots, r_k\}$\\}

\begin{enumerate}
    \item Solve it with Q-Learning. gamma=0.9
    \item Transitions from one state to another are only allowed based on provided 'Actions'.
    Transitions from one state to another are only allowed based on possible actions. 
    \item First, simulate the environment for 1000 episodes and fill in the Q-table for states. If an episode goes more than 100 steps terminate that episode.
    \item Episodes MUST begin at "Start" and finish at "End".
    \item Print the state/action pairs and their "Q-values" for the optimal episode from "Start" to "End".  
    
\end{enumerate}
\end{tcolorbox}\vspace{-3mm}
\caption{This prompt is used for iterative verification of the LLM outputs until reaching the desired outputs. \textbf{Task}, \textbf{States}, \textbf{Actions}, and \textbf{Rewards} are place-holders for use case specific values, e.g., values from Figure~\ref{prompt_all} can be used here for research scientist workflow.}
\label{prompt_iterative}
\end{figure*}

\section{Prompts}
Figures~\ref{prompt_all} and~\ref{prompt_iterative} compile all the prompts used for our demonstrated use cases. 
Because of the generalizability in their designs, these prompts can readily serve as a template for formulating additional RL-based use cases.
In addition to leveraging self-reflection~\cite{shinn2023reflexion}, we also implemented capitalization of critical instructions to emphasize the output requirements and ensure more consistent generations from the LLM~\cite{kang2023llm}. 
The aforementioned prompting techniques, in addition to the code generation capabilities of GPT~\cite{OpenAI2023PlugIns}, allow the LLM to accurately implement and simulate the MDP based on the provided input and thereby produce the optimal policy.

\end{document}